\pdfoutput=1
\documentclass[11pt]{article}

\usepackage[final]{acl}

\usepackage{times}
\usepackage{latexsym}

\usepackage[T1]{fontenc}
\usepackage[utf8]{inputenc}

\usepackage{microtype}

\usepackage{graphicx}
\usepackage{booktabs}
\usepackage{inconsolata}
\usepackage{amsmath}
\usepackage{bm}
\usepackage{multirow}
\usepackage[normalem]{ulem}
\useunder{\uline}{\ul}{}

\usepackage{bbding}
\usepackage{color}
\usepackage{subfigure}

\usepackage{colortbl}
\usepackage{hyperref}

\newcommand{\model}{{KnowPAT}}
\newcommand{\improve}[1]{{\textbf{\color{red}#1}}}
\newcommand{\rmnum}[1]{\romannumeral #1}
\newcommand{\Rmnum}[1]{\expandafter\@slowromancap\romannumeral #1@}

\definecolor{light-gray}{gray}{0.96}
\title{Knowledgeable Preference Alignment for LLMs in \\ Domain-specific Question Answering}

\author{
    Yichi Zhang$^\spadesuit$$^\diamondsuit$\footnotemark[1],
    Zhuo Chen$^\spadesuit$$^\diamondsuit$\thanks{~~Equal contribution.},
    Yin Fang$^\spadesuit$$^\diamondsuit$, 
    \textbf{Yanxi Lu}$^\clubsuit$,
    \textbf{Fangming Li}$^\clubsuit$,\\
    \textbf{Wen Zhang}$^\spadesuit$$^\diamondsuit$\footnotemark[2],
    \textbf{Huajun Chen}$^\spadesuit$$^\diamondsuit$\thanks{~~Corresponding authors.}\\
    $^\spadesuit$ Zhejiang University \\
    $^\diamondsuit$ Zhejiang University - Ant Group Joint Laboratory of Knowledge Graph\\
    $^\clubsuit$ NAIE, Huawei Technologies Co.Ltd.\\
    \texttt{
    \{zhangyichi2022,zhuo.chen,zhang.wen,huajunsir\}@zju.edu.cn 
    }\\
  \url{https://github.com/zjukg/KnowPAT}
}

\begin{document}

\maketitle

\begin{abstract}
Deploying large language models (LLMs) to real scenarios for domain-specific question answering (QA) is a key thrust for LLM applications, which poses numerous challenges, especially in ensuring that responses are both accommodating to user requirements and appropriately leveraging domain-specific knowledge bases. They are the two major difficulties for LLM application as vanilla fine-tuning falls short of addressing. Combining these requirements, we conceive of them as the requirement for the model's \textbf{preference} to be harmoniously aligned with humans'. Thus, we introduce \textbf{Know}ledgeable \textbf{P}reference \textbf{A}lignmen\textbf{T} ({\model}), which constructs two kinds of preference sets to tackle the two issues.  Besides, we design a new alignment objective to align the LLM preference with different human preferences uniformly, aiming to optimize LLM performance in \textbf{real-world, domain-specific} QA settings. Adequate experiments and comprehensive comparisons with 15 baseline methods illustrate that our {\model} is a superior pipeline for real-scenario domain-specific QA with LLMs. % Our code is available at \href{https://anonymous.4open.science/r/KnowPAT-0135}{this anonymous github link}.
\end{abstract}

\vspace{-8pt}
\section{Introduction}

\begin{figure}[]
  \centering
\includegraphics[width=0.9\linewidth]{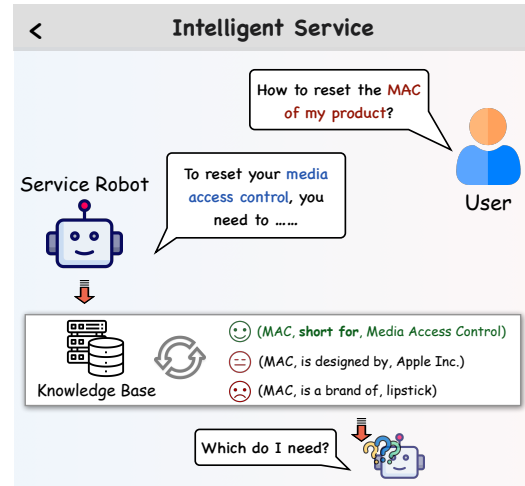}
\vspace{-8pt}
  \caption{A simple case of intelligent service for cloud products. Such a simple example is meant to illustrate the importance of selective use of retrieved knowledge as MAC is a terminology in computer networking rather than a kind of computer or lipstick in the user context.}
  \label{introduction}
  \vspace{-16pt}
\end{figure}
In contemporary digital commerce platforms, the deployment of automated and intelligent \textbf{question-answering} (QA) services is a pivotal task to augment service quality. These services are designed to furnish answers to domain-specific customer queries. Building such a domain-specific QA system, while highly sought after, remains a daunting challenge in practical scenarios.

\par Domain-specific QA necessitates a comprehensive understanding of a specific domain to answer specialized questions. However, traditional deep learning models \cite{DBLP:conf/naacl/BERT, DBLP:journals/jmlr/T5} still have insufficient domain-specific expertise. This makes the domain \textbf{knowledge bases} (KBs) \cite{liang2022kgsurvey} a pivotal tool for the storage and querying of domain knowledge. KBs can store human knowledge in the triple form, offering a unified, maintainable, and extensible representation of the knowledge from heterogeneous sources. The utility of KBs has already been demonstrated across various application scenarios such as E-commerce \cite{DBLP:conf/mm/mmkg-ecom}, and health care \cite{DBLP:journals/artmed/mmkghc1}. Within the context of QA, incorporating external knowledge source represents a promising approach, which is known as KBQA \cite{DBLP:conf/iclr/JiangZ0W23}.

\par Meanwhile, as large language models (LLMs) \cite{west2023generative} achieve significant progress and exhibit substantial proficiency within numerous NLP fields \cite{DBLP:journals/corr/llm4kgc-zhu}, applying LLMs into various downstream tasks have been a predominant trend in industry \cite{DBLP:conf/recsys/llm4rec1}. Contrasting earlier pre-trained language models \cite{DBLP:conf/naacl/BERT, DBLP:journals/jmlr/T5}, LLMs trained on the massive corpus have outperformed text generation capabilities which perform better when interacting with human \cite{DBLP:conf/nips/instructgpt}. To adapt the LLMs for downstream usage, supervised fine-tuning (SFT) \cite{DBLP:journals/corr/tuning-survey} is applied to fit the model with specific tasks and data. However, the LLM application for real-scenario QA with external KB remains an underexplored domain, with limited work addressing this intersection.

\par Our goal entails the resolution of a challenge in real-world applications: \textbf{How can LLM be used to solve real-scenario QA problems supported by external knowledge bases?} A generic pipeline for this problem is the retrieve-augmented generation (RAG) \cite{DBLP:journals/corr/kbqa2}, which first retrieves relative knowledge triples for the question as reference data and subsequently fine-tunes the LLM with knowledge-enhanced prompt. However, this conventional approach often encounters obstacles in practical scenarios. Firstly, the LLM-produced responses must prioritize user-friendliness, avoiding any generation of inappropriate or unfriendly content. Secondly, the retrieved knowledge is not invariably useful, necessitating that LLMs develop the capacity to judiciously exploit knowledge. Figure \ref{introduction} illustrates a simple case in which retrieved knowledge is not always desperately needed (e.g., MAC is a kind of lipstick), which requires the LLMs to selectively utilize the retrieved knowledge instead of generating answers without thoughtful consideration. These two issues can uniformly collectively constitute the preference problem of LLMs. LLMs have their \textbf{style preference} to generate contents and \textbf{knowledge preference} to selectively use the retrieved knowledge in the prompt. \textbf{As a practical application, the preference of LLMs needs to align with human expectations and requirements for better service}. This refers to preference alignment (PA) \cite{DBLP:journals/corr/rrhf}, a burgeoning topic in the LLMs community, which would incorporate human preference to tune the LLMs during training. PA aims to control the model to generate human-preferred content and avoid unpreferred content. However, the scenarios faced by current PA works tend to be generic. No research has been explicitly directed towards domain-specific applications such as our scenario, providing impetus for further exploration.
\par In this paper, we propose a novel three-step \textbf{Know}ledgeable \textbf{P}reference \textbf{A}lignmen\textbf{T} ({\model}) pipeline to address the domain-specific QA task for a real-scenario LLM application. {\model} propose \textbf{knowledgeable preference set construction} to incorporate domain KBs to construct knowledgeable preference data. Besides, a new alignment objective is designed to optimize the LLM with the knowledge preference.
Our contribution can be summarized as three-folded:

\noindent (1). We are the first work that introduces preference alignment for domain-specific QA with LLMs and domain KBs, which is an industrial practice with practical applications.

\noindent (2). We propose a knowledgeable preference alignment (\model) framework to incorporate KBs into the preference alignment process of LLMs. We balanced the need for both style and knowledge preference and devised a new training objective to align the LLM with human preference.

\noindent (3). We conduct comprehensive experiments to validate the effectiveness of our methods on two datasets, which shows that {\model} outperforms 15 existing baselines.

\section{Problem Setting}
In this section, we will first introduce our problem scenario and basic notations.

\begin{figure*}[]
  \centering
\includegraphics[width=\linewidth]{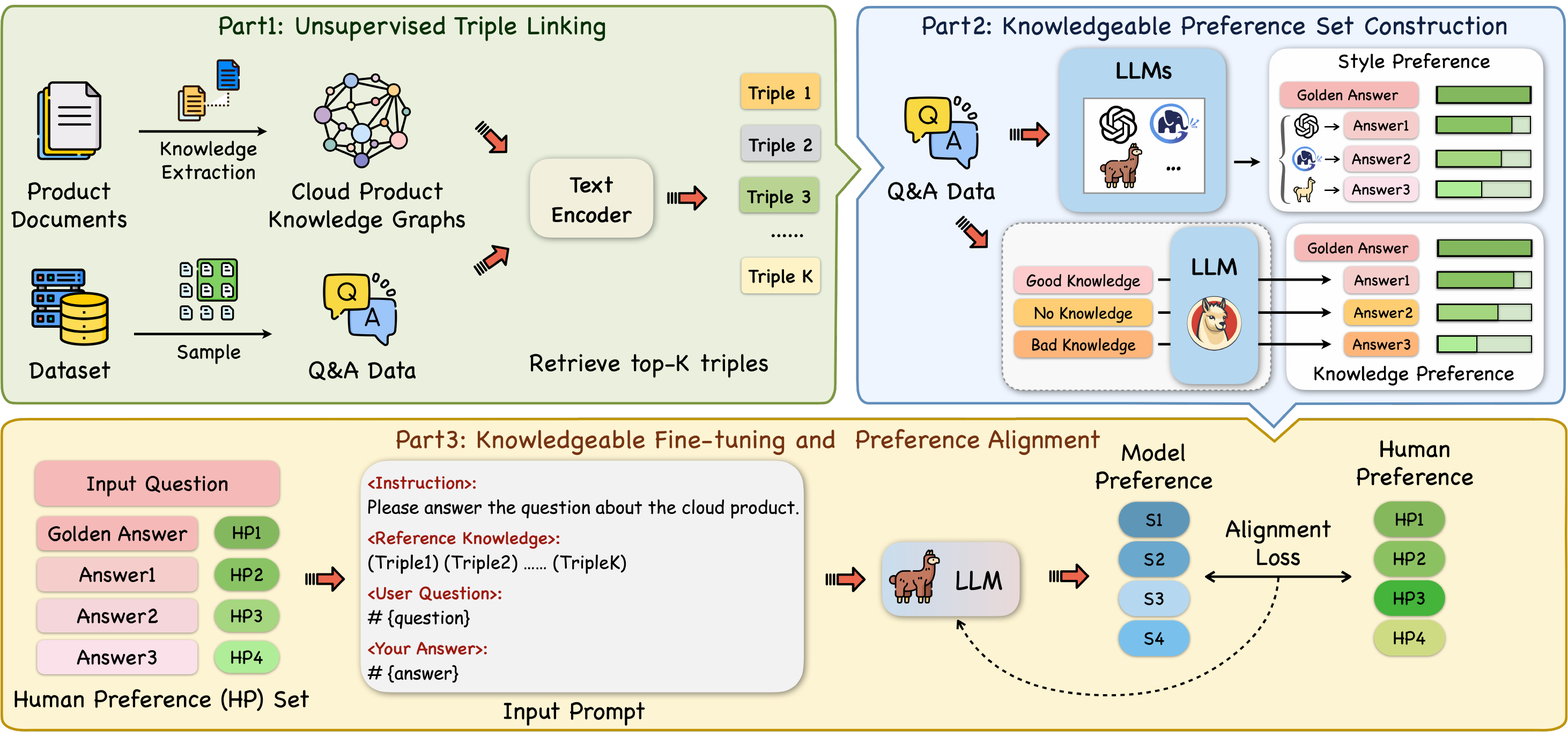}
  \caption{The overall architecture of {\model}. We design three important modules in our framework: unsupervised knowledge retrieve, knowledgeable preference set construction, and knowledgeable preference alignment.} %We first retrieve relative knowledge triples for the question in part 1 and apply the retrieved knowledge to construct the knowledgeable preference set in part 2. The preference sets will participate in the fine-tuning and preference alignment process in part 3, which will align the LLM with human preference.}
  \label{model}
  \vspace{-16pt}
\end{figure*}

Our overall target is to fine-tune a LLM $\mathcal{M}$ with our QA datasets $\mathcal{D}=\{(q_i, a_i)\mid i=1,2,\dots, N\}$ where $q_i, a_i$ represent a question and answer pair. The questions in the dataset are all about common usage issues with our cloud products while the questions and answers are manually collected and labeled, which are gloden answers  with decent and knowledgeable responses. For vanilla fine-tuning (VFT), we first wrap the QA pair with a prompt template $\mathcal{I}$ and the model $\mathcal{M}$ is autogressively \cite{DBLP:conf/nips/gpt3} optimized as:

\begin{small}
\begin{equation}
\label{vft}
    \mathcal{L}_{ft}=-\frac{1}{|a_i|}\sum_{j=1}^{|a_i|}\log P_{\mathcal{M}}(a_{i, j}|\mathcal{I}, q_i, a_{i, <j})
\end{equation}
\end{small}
where $a_{i, j}$ is the $j$-th token of $a_i$ and $P_{\mathcal{M}}$ denotes the token probability predicted by the model $\mathcal{M}$. With such a training objective, the training QA data serves as the supervision information to tune the model $\mathcal{M}$ to the QA scenario. Besides, as a domain-specific task, we maintain a \textbf{domain knowledge base} (domain KB) $\mathcal{B}$. The domain KB can take very many forms, such as domain knowledge graphs, or documents. We denote $k_i\in\mathcal{B}$ as one support knowledge in the domain KB.
%The CPKG is denoted as $\mathcal{G}=(\mathcal{E},\mathcal{R},\mathcal{T})$ where $\mathcal{E},\mathcal{R},\mathcal{T}$ are the entity set, relation set, and triple set respectively. The knowledge graph will be used as an external knowledge source to support the model for QA. 
By retrieving top-k knowledge $k$ with higher relevance, the input prompt will incorporate the retrieved knowledge $\mathcal{K}$. Thus, $\mathcal{M}$ can learn the relative knowledge during the VFT process, which is a general retrieve-augmented generation (RAG) pipeline for domain-specific LLM applications.

\par However, such a VFT approach can not achieve pretty good results for the domain-specific QA. On the one hand, applications in real scenarios should be user-friendly, otherwise, they will not bring commercial value. Thus, the text style of the generated response should be more acceptable for users. On the other hand, the knowledge retrieval process is unsupervised and the effectiveness of the retrieved knowledge is hard to guarantee, which means that the model $\mathcal{M}$ needs to acquire the ability to judge and selectively utilize the knowledge triples.
Therefore, we should improve the basic VFT to solve these two problems. 

\par Actually, both of these problems can be summarised as model preference. The LLM $\mathcal{M}$ has its style preference to generate texts and its knowledge preference to selectively utilize the retrieved knowledge. For the model to be practically applicable, the model preference should align with human preference, aiming to generate high-quality answers that humans prefer. Preference alignment (PA) is an important topic for LLMs. To apply PA during LLM fine-tuning, we sample a preference set $\mathcal{P}=\{b_1, b_2, \dots, b_l\}$ with $l$ different answers for each QA pair $(q, a)$. We denote $r_i$ as the preference score of each answer $b_i$ where higher $r_i$ represents that humans prefer this answer. During training, we will define another objective $\mathcal{L}_{align}$ to align the model $\mathcal{M}$ with the preference set $\mathcal{P}$, aiming to increase the probability of a human preferred answer appearing and simultaneously decrease the probability of an unpreferred answer. The human preference of each answer is the preference score $r$. The overall training objective then becomes $\mathcal{L}=\mathcal{L}_{ft}+\mathcal{L}_{align}$.With such a multi-task objective, the LLM is fine-tuned to \textbf{fit the golden answers while avoiding unpreferred results}. The next question is how to generate a preference set to reflect both the style and knowledge preference.

\section{Our KnowPAT Pipeline}
In this section, we will present our pipeline of knowledgeable preference alignment ({\model}), which consists of three key parts: unsupervised knowledge retrieve, knowledgeable preference set construction, fine-tuning, and training. Figure \ref{model} demonstrates an intuitive view of the three parts in our pipeline design.
\subsection{Unsupervised Knowledge Retrieve}
The first key parts is the unsupervised knowledge retrieve which aims to link the knowledge in the KB $\mathcal{B}$ to each question $q_i$. We design a simple semantic similarity-based retriever $\mathcal{H}$ to achieve this goal. The similarity between the $i$-th question $q_i$ and the $j$-th knowledge $k_j$ is:

\begin{small}
\begin{equation}
    sim(i, j)=\mathbf{Cosine}(\mathcal{H}(q_i), \mathcal{H}(k_j))
\end{equation}
\end{small}
where the retriever $\mathcal{H}$ serves as a textual encoder and we treat both the question and knowledge as a text sequence to get their sentence representations. The similarity is based on the cosine similarity of the two representations. We retrieve the top-$k$ knowledge with the highest similarities for each question $q_i$ and denote the \textbf{retrieved knowledge} (RK) as $\mathcal{K}$. RK will be added into the input prompt as the background knowledge for the current question.
\par This process is unsupervised as we have no manually labeled question-knowledge pairs. Besides, our model will be deployed for real scenario usage, so it also requires strong zero-shot generalization capabilities to new questions. For these two reasons, the retrieved knowledge $\mathcal{K}$ might be noisy and useless to provide background knowledge. We think that the LLM $\mathcal{M}$ should learn the knowledge preference to select helpful information from the retrieved knowledge $\mathcal{K}$.

\subsection{Knowledgeable Preference Set Construction}
Motivated by such goal, we propose a knowledgeable preference set construction process to enable retrieved knowledge in the preference set construction, which consists of two parts: the style and the knowledge preference set.

% 强调专家的偏好，和榜单的结果相结合。
\par For the \textbf{style preference set (SPS)} $\mathcal{P}_s$, we select $l-1$ different LLMs denoted $\mathcal{M}_1, \mathcal{M}_2, \dots, \mathcal{M}_{l-1}$. These different LLMs $\mathcal{M}_{i}$ have different textual comprehension and expression skills, which can generate answers with different text styles. The ability and quality of these models to answer domain-specific questions are inferior compared to human-labeled golden answers. The $l-1$ answers generated in this way and golden answers form a style preference set $\mathcal{P}_{s}=\{b_1, b_2, \dots, b_l\}$ with length $l$. For the \textbf{knowledge preference set (KPS)}, we assume that the knowledge that has high similarity but do not reach the top-$k$ rank are more likely to be knowledge that is not useful for the input question. We can get preference sets with different quality by retrieving some relatively worse knowledge and prompting the model to generate responses with knowledge of different quality. In our design, we retrieve 3 groups of knowledge $\mathcal{K}_{1},\mathcal{K}_{2},\mathcal{K}_{3}$ from the CPKG. $\mathcal{K}_{1}$ represents the retrieved top-$k$ knowledge, $\mathcal{K}_2=\emptyset$ is an empty set with no retrieved knowledge. $\mathcal{K}_{3}$ represents the knowledge with top $k+1$ to $2k$ similarities which we think are easily misused knowledge with relatively high semantic similarity. Then we wrap the different knowledge $\mathcal{K}_{i}$ with the input prompt $\mathcal{I}$ into the LLM $\mathcal{M}$ and generate different answers. These generated 3 answers and the golden answer form a knowledge preference set $\mathcal{P}_{k}=\{c_1, c_2, c_3, c_4\}$.

\par By doing this, we can get two preference sets for each QA pair. To simplify the setting, we set $l=4$ to let the two sets be of the same size. Besides, we design a rule-based strategy to decide the preference score $r$ for each answer. For the style preference set $\mathcal{P}_s$, the high-quality golden answer $b_1$ is assigned with the highest score, and answers from other LLMs were determined by their general capabilities. In practice, we choose three different LLMs ChatGPT ($b_2$) \cite{DBLP:conf/nips/instructgpt}, ChatGLM-6B ($b_3$) \cite{DBLP:conf/iclr/chatglm}, and Vicuna-7B ($b_4$). The results of several LLM ranking lists indicate that the three are ranked in order of ability as follows ChatGPT > ChatGLM > Vicuna. Besides, after verification by human experts, we also believe that the quality of the answers generated by these three models in our QA scenarios also conforms to this rule.  Thus, the preference scores are assigned in this order: $r_1>r_2>r_3>r_4$. 

\par Meanwhile, for the knowledge preference set $\mathcal{P}_k$, the golden answer $c_1$ still has the highest preference score $r_1$. The answer $c_2$ generated with top-$k$ knowledge $\mathcal{K}_1$ has the second highest preference. The answer $c_3$ generated with no extra knowledge $\mathcal{K}_2$ has the third highest preference, and the answer $c_4$ generated with knowledge $\mathcal{K}_3$ is the worst. We found in our actual tests that the mismatch rate between the retrieved knowledge $\mathcal{K}_3$ and the question $q$ is very high and easily misleads the model $\mathcal{M}$, so we set its score to be lower than the case of the empty knowledge $\mathcal{K}_2$. Thus, for the knowledge preference set $\mathcal{P}_{k}$, the preference scores are still in the order: $r_1>r_2>r_3>r_4$. For each QA pair, we can construct two preference sets and we finally get the whole preference data with $2N$ preference sets. The preference data will participate in the fine-tuning process to control the style preference and knowledge preference for the model $\mathcal{M}$. Note that the size of the two preference sets need not be strictly same, and we have adopted the above formulation for the sake of uniformity of representation in our paper.

\subsection{Fine-tuning and Preference Alignment}
\par In addition to the vanilla fine-tuning loss $\mathcal{L}_{ft}$ with the golden answer, the preference data will also participate in the training process. For each preference set, the preference score $r_i$ of the i-th answer represents our degree of preference. We expect the model $\mathcal{M}$ to align with our preference. Thus, we design another score to represent the preference of the model, which is denoted as:

\begin{small}
\begin{equation}
\label{score}
    \mathcal{S}_{i}=\frac{1}{|a_i|}\sum_{j=1}^{|a_i|}\log P_{\mathcal{M}}(a_{i, j}|\mathcal{I}, q_i, a_{i, <j})
\end{equation}
\end{small}

\par This score $\mathcal{S}_i$ is the average log-likelihood of each answer token conditioned on the given prompt template $\mathcal{I}$ and question $q_i$. Higher scores represent a higher probability that the model considers the current answer to occur. To align the model preference with our envision, we designed a new alignment objective for our scenario. The alignment objective is denoted as:

\begin{small}
\begin{equation}
\begin{aligned}
    \mathcal{L}_{align}=-\sum_{i=1}^{|\mathcal{P}| -1}\left(\log\sigma(\mathcal{S}_i)+\sum_{r_j<r_i}\log\sigma(-\mathcal{S}_j)\right)\\
    % =&\sum_{i=1}^{|\mathcal{P}| -1} \left(\log(1+e^{-\mathcal{S}_i})+\sum_{r_j<r_i}\log(1+e^{\mathcal{S}_j})\right)
\end{aligned}
\end{equation} 
\end{small}

where $\sigma$ is the sigmoid function. Such an objective is newly proposed by us to achieve the preference alignment process, which contrasts the preferred answer and the unpreferred answers. It is worth noting that the human preference scores $r_i$ will only determine the ordering corresponding to different answers and will not be directly involved in the computation and gradient accumulation. Existing methods like RRHF \cite{DBLP:journals/corr/rrhf} and SLiC-HF \cite{DBLP:journals/corr/slic-hf} apply a margin-rank loss in the form $\sum_{r_j<r_i}\max(0, \lambda-\mathcal{S}_{i}+\mathcal{S}_{j})$ to achieve preference alignment. But their design only optimizes the model preference when the model preference score $\mathcal{S}$ of a human preferred answer is lower than an unpreferred answer (a more formalized formulation would be $\mathcal{S}_i<\mathcal{S}_j$ when $r_j<r_i$). However, we think that the preference should still be optimized in this situation and propose such a training objective to continuously decrease the occurrence probability of the unpreferred answers. Meanwhile, as different answers have different text quality and preference degrees, we further design an adaptive weight to control the influence of each preferred answer, which is denoted as:

\begin{small}
\begin{equation}
    \mu_{i}=\frac{\mathcal{S}_{i}-\mathcal{S}_{min}}{\mathcal{S}_{max}-\mathcal{S}_{min}}
\end{equation}
\end{small}
where $\mathcal{S}_{max}$ and $\mathcal{S}_{min}$ are the max and min model preference scores in a preference set $\mathcal{P}$. With such an adaptive weight, the influence of the answers with different preferences could be dynamically adjusted. The alignment loss then becomes:

\begin{small}
\begin{equation}
\mathcal{L}_{align}=\sum_{i=1}^{|\mathcal{P}| -1}\mu_{i} \left(\log(1+e^{-\mathcal{S}_i})+\sum_{r_j<r_i}\log(1+e^{\mathcal{S}_j})\right)
\end{equation}
\end{small}

\par The final training objective is still in a multi-task manner and we add a hyper-parameter $\lambda$ as the coefficient of the alignment loss:

\begin{small}
\begin{equation}
\mathcal{L}=\mathcal{L}_{ft}+\frac{\lambda}{|\mathcal{P}| - 1}\mathcal{L}_{align}
\end{equation}
\end{small}
where $\mathcal{P} - 1$ represents the count of prefer-unprefer contrast to normalize the alignment loss. For each preference set constructed in the previous section, the model is trained and optimized with such an objective. % Note that the size of each preference set does not have to be 4, the hyperparameter we used in the previous design to better fit the theme of knowledgeable preferences.

%\subsection{Model Inference and Deployment}
%After training the model $\mathcal{M}$ with the multi-task objective, the model would be used for further validation and application. The inference process of $\mathcal{M}$ can be denoted as:
%\begin{equation}
%\label{inference}
%a=\arg\max_{a}P_{\mathcal{M}}(a|\mathcal{I},\mathcal{K},q)
%\end{equation}
%where $\mathcal{I},\mathcal{K},q$ are the prompt template, retrived knowledge, and the question respectively. But as an application that needs to be used in real life, the model preference should \textbf{align with human preference}. 

% 这里接一段部署相关的描述，问华为的人要。
\section{Experiments and Analysis}

\begin{table*}[]
\caption{The experimental results for traditional text generation metrics on two datasets. The \improve{red} numbers represent the improvement of {\model} on each dataset. The best baseline performance is \underline{underlined}.}
\label{main-exp}
\centering
\resizebox{0.9\textwidth}{!}{
\begin{tabular}{c|cc|cccccccc}
\toprule
\textbf{Dataset} & \textbf{Type} & \textbf{Setting} & \textbf{BLEU-1} & \textbf{BLEU-2} & \textbf{BLEU-3} & \textbf{BLEU-4} & \textbf{ROUGE-1} & \textbf{ROUGE-2} & \textbf{ROUGE-L} & \textbf{METEOR} \\
\midrule
\multirow{18}{*}{\rotatebox{90}{\textbf{CPQA}}} & \multirow{5}{*}{Zero-shot} & Vicuna & 14.18 & 7.89 & 5.02 & 2.69 & 16.31 & 6.15 & 15.69 & 17.96 \\
 &  & ChatGLM & 14.21 & 8.36 & 5.41 & 2.79 & 15.38 & 5.64 & 14.75 & 19.34 \\
 &  & Baichuan & 15.51 & 9.08 & 5.86 & 2.87 & 16.74 & 6.64 & 15.81 & 19.71 \\
 &  & Atom & 10.07 & 4.11 & 2.06 & 8.15 & 6.24 & 1.99 & 6.02 & 11.31 \\
 &  & ChatGPT & 13.09 & 7.72 & 4.93 & 2.59 & 16.96 & 6.68 & 16.15 & 19.52 \\
 \cmidrule{2-11} 
 & \multirow{4}{*}{\begin{tabular}[c]{@{}c@{}}In-context\\ Learning\end{tabular}} & 1-shot & 8.97 & 3.84 & 1.88 & 0.53 & 7.49 & 1.99 & 7.31 & 10.41 \\
 &  & 2-shot & 9.11 & 3.84 & 1.85 & 0.50 & 7.34 & 1.82 & 7.01 & 9.88 \\
 &  & 4-shot & 8.18 & 3.42 & 1.65 & 0.48 & 7.07 & 2.04 & 6.91 & 8.83 \\
 &  & 8-shot & 7.79 & 3.29 & 1.70 & 0.79 & 6.57 & 1.38 & 6.41 & 8.19 \\ \cmidrule{2-11} 
 & Fine-tuning & Atom & 14.89 & 9.35 & 7.33 & 6.05 & 14.77 & 5.57 & 14.61 & 15.99 \\
 \cmidrule{2-11} 
 & \multirow{6}{*}{Alignment} & DPO & 18.31 & 12.07 & 9.63 & 7.81 & 17.74 & 6.61 & 17.38 & 18.81 \\
 & & RRHF & 11.99 & 6.32 & 4.52 & 3.47 & 12.56 & 4.08 & 12.29 & 12.62 \\
 
 &  & SLiC & 16.55 & 10.34 & 7.99 & 6.53 & 14.69 & 5.03 & 14.48 & 16.95 \\
 &  & PRO & 18.27 & \underline{12.36} & \underline{10.04} & \underline{8.41} & 17.07 & 6.75 & 16.85 & 19.17 \\
 &  & AFT-BC & \underline{18.39} & 12.17 & 9.86 & 7.81 & \underline{18.09} & \underline{7.14} & \underline{17.76} & \underline{19.48} \\
 &  & AFT-DC & 15.34 & 8.44 & 5.94 & 4.35 & 14.51 & 5.59 & 14.15 & 16.31 \\
 \cmidrule{2-11} 
 & \multicolumn{2}{c|}{\multirow{2}{*}{\textsc{KnowPAT}}} & 21.87 & 15.59 & 13.21 & 11.14 & 19.91 & 8.31 & 19.62 & 22.42 \\
 & \multicolumn{2}{c|}{} & \improve{+18.92\%} & \improve{+26.13\%} & \improve{+31.57\%} & \improve{+32.46\%} & \improve{+10.06\%} & \improve{+16.38\%} & \improve{+10.47\%} & \improve{+15.09\%} \\
\midrule
& Fine-tuning & Atom & 23.12 & 15.17 & 10.89 & 8.31 & 7.92 & 0.79 & 7.17 & 21.26 \\
\cmidrule{2-11}
\multirow{8}{*}{\rotatebox{90}{\textbf{RJUA-QA}}} & \multirow{7}{*}{Alignment} & DPO & 23.87 & 15.81 & 11.67 & 8.99 & 11.69 & \underline{2.53} & 9.66 & 22.43 \\
 &  & RRHF & 22.32 & 14.94 & 10.86 & 8.41 & 7.39 & 1.40 & 5.92 & 21.66 \\
 &  & SLiC & 23.51 & 15.46 & 11.27 & 8.69 & 8.68 & 1.17 & 7.70 & 22.14 \\
 &  & PRO & 24.01 & 16.05 & 11.76 & 9.06 & \underline{12.50} & 1.69 & \underline{9.98} & 22.38 \\
 &  & AFT-BC & \underline{24.43} & \underline{16.27} & \underline{12.10} & \underline{9.37} & 9.06 & 2.03 & 7.31 & \underline{23.30} \\
 &  & AFT-DC & 20.81 & 13.01 & 9.15 & 6.92 & 6.40 & 0.65 & 5.08 & 20.02 \\
 \cmidrule{2-11} 
 & \multicolumn{2}{c|}{\multirow{2}{*}{\textsc{KnowPAT}}} & 25.61 & 18.04 & 13.95 & 11.38 & 10.75 & 4.26 & 10.46 & 24.48 \\
 & \multicolumn{2}{c|}{} & \improve{+4.83\%} & \improve{+10.88\%} & \improve{+15.28\%} & \improve{+21.45\%} & \improve{-} & \improve{+68.38\%} & \improve{+4.82\%} & \improve{+5.64\%} \\
 \bottomrule
\end{tabular}
}
\end{table*}

\begin{table}[]
\caption{The experimental results of model-based metrics. We report the BERTScore, reward score, and perplexity (PPL) for KnowPAT and the baseline methods. The best result of each metric is bold and the second best is \underline{underlined}.}
\vspace{-8pt}
\label{main-exp2}
\centering
\resizebox{0.8\columnwidth}{!}{
\begin{tabular}{c|ccc}
\toprule
  & \textbf{BERTScore$\uparrow$} & \textbf{Reward$\uparrow$} & \textbf{PPL$\downarrow$} \\
\midrule
Fine Tuning & 66.24  & \textbf{-1.64} & 31.13  \\
RRHF  & 64.48  & \underline{-1.67} & 31.26  \\
SLiC  & 66.69  & -1.74 & 32.51  \\
PRO & \underline{67.41}  & -1.78 & 32.37  \\
AFT & 66.16  & -2.25 & \underline{30.11}  \\
\midrule
\textbf{KnowPAT} & \textbf{69.34}  & -1.69  & \textbf{29.93} \\
\bottomrule
\end{tabular}
}
\vspace{-16pt}
\end{table}

In this section, we present the detailed experimental settings and analyze the experiment results to investigate the following four research questions:

\noindent \textbf{{(\rmnum{1})}} \textbf{RQ1:} How does {\model} perform compared with the baseline methods?

\noindent \textbf{{(\rmnum{2})}} \textbf{RQ2:} Do the proposed modules in {\model} really benefit the performance of {\model}?

\noindent \textbf{{(\rmnum{3})}} \textbf{RQ3:} Are there some intuitive cases to demonstrate the effectiveness of {\model}.

\noindent \textbf{{(\rmnum{4})}} \textbf{RQ4:} Does the LLM still keep the general ability rather than catastrophic forgetting?

\par These four questions evaluate our approach on four dimensions: performance, design soundness, intuition, and usability in real scenarios. We will answer the four questions in the following sections.
\subsection{Experiment Settings}
\subsubsection{Dataset Information}
Our experiments are performed on both private and public datasets. The private CPQA dataset consists of a CPKG and QA pairs. The public dataset is RJUA-QA \cite{RJUA}, which is a urology domain QA dataset. The detailed dataset information is presented in Appendix \ref{appendix::dataset}. % The first part is the CPKG with 13995 entities, 463 relations, and 20752 triples. The second part is the QA dataset with 8909 QA pairs. For each data instance in the training, we construct two preference sets and  get 15818 preference sets with 4 answers in each set.

\subsubsection{Baseline Methods}
To make a comprehensive study, we select four types of different baseline methods to demonstrate the effectiveness of our preference alignment approach. We not only want to show that alignment is a better framework for LLM application compared to other paradigms (e.g. zero-shot reasoning, in-context learning \cite{DBLP:journals/corr/icl-survey}, vanilla fine-tuning \cite{DBLP:conf/nips/instructgpt,DBLP:journals/corr/abs-2306-08018}), but also to show that our method is better than other preference alignment methods \cite{DBLP:journals/corr/rrhf, DBLP:journals/corr/slic-hf, DBLP:journals/corr/pro,DBLP:journals/corr/AFT, DBLP:journals/corr/dpo}. The detailed information of the baselines are shown in Appendix \ref{appendix::baseline}.

\subsubsection{Evaluation Metrics}
To make a comprehensive evaluation of the experimental results, we employ the different evaluation metrics from three aspects: traditional text generation metrics (BLEU \cite{DBLP:conf/acl/BLEU}, ROUGE \cite{lin2004rouge}, CIDEr \cite{DBLP:conf/cvpr/CIDEr}, and METEOR \cite{DBLP:conf/acl/METEOR}), model-based metrics (BERTScore \cite{DBLP:conf/iclr/BERTSCORE}, PPL), and manual evaluation. The detailed information of the evaluation metrics refers to Appendix \ref{appendix::evaluation}.

\subsubsection{Implementation Details}

In our experiment, we select Atom-7B \footnote{\url{https://github.com/FlagAlpha/Llama2-Chinese}} as the backbone LLM $\mathcal{M}$, which is an open-source version of Llama2 \cite{DBLP:journals/corr/llama2, DBLP:journals/corr/llama} with Chinese vocabulary extension. As our dataset is mainly in Chinese, we choose Atom-7B-chat to be our backbone model for experiments. Another consideration for us is that using the open-source Llama architecture model enhances the generality of our method to maintain the community ecology of LLMs.
% Meanwhile, the Atom-7B-chat is first fine-tuned on our cloud product documents which is a domain corpus to enable the backbone model with a basic understanding of the domain concepts. We name this model Atom-7B-CP and employ this fine-tuned version as our backbone. 
For unsupervised triple linking, BGE-base-zh-v1.5 \cite{bge_embedding} is applied as the retriever $\mathcal{H}$ to encode and retrieve relative knowledge candidates.

\par During training, we tune the backbone model with bf16 float precision. The training epoch is set to 3 and the gradient accumulation step is set to 8. We optimize the model using AdamW optimizer \cite{DBLP:conf/iclr/adamw} while the learning rate is fixed to $3e^{-4}$. The coefficient hyper-parameter $\lambda$ is search in $\{1,0.1,0.01,0.001\}$.

\subsection{Main Results (Q1)}
The main results of the traditional metrics are shown in Table \ref{main-exp}. As we mentioned before, the traditional metrics can measure the similarity between the generated answer and the golden answer. From the results, we can observe that {\model} achieved obvious improvements compared with the baseline methods. We can conclude that {\model} achieves a more significant improvement in the BLEU-3(42.03\%)/BLEU-4(43.99\%) than BLEU-1(22.67\%)/BLEU-2(34.79\%), which means that {\model} makes more significant progress in capturing some complex phrases and discourse. Corresponding to our cloud product QA scenario, these complex phrase usages are usually specialized terms that have a critical impact on the quality of the answer.

\par Besides, we evaluate our methods with three model-based metrics BERTScore \cite{DBLP:conf/iclr/BERTSCORE}, reward score \cite{DBLP:journals/corr/rrhf}, and PPL \cite{DBLP:journals/corr/rrhf}, which is shown in Table \ref{main-exp2}. We can observe that {\model} still achieves good performance in the model-based metrics such as BERTScore and PPL, which means that the results generated by {\model} are more acceptable for the language models. For the reward score, relatively good results have also been achieved by
{\model}.

\par
\begin{figure}[]
  \centering
\includegraphics[trim=28 0 0 0,width=0.9\linewidth]{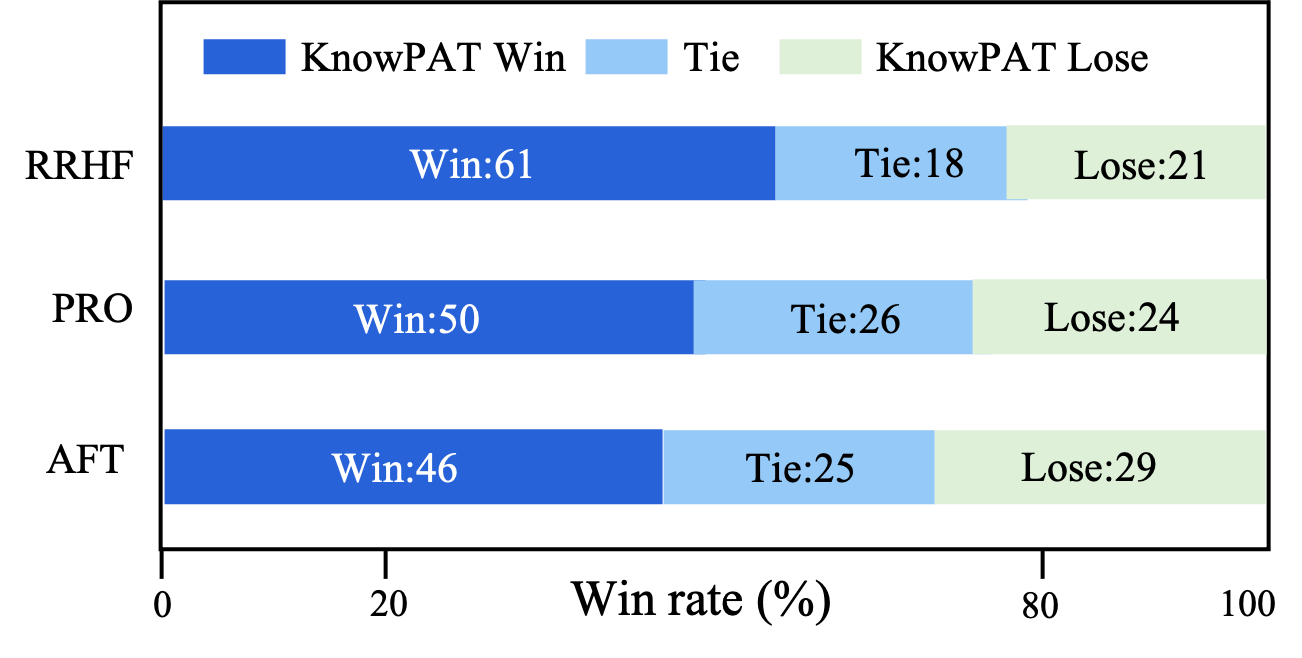}
\vspace{-8pt}
  \caption{The human evaluation results. For each competition, we randomly select 100 questions and compare the generated results of the two methods.}
  \label{human-eval}
\vspace{-10pt}
\end{figure}

 Further, we conduct a human evaluation for our method and baseline methods. The two results from the two models are shown to the human evaluator anonymously so that the human evaluator can choose a better result. The model which generates that result will get one point and the competition results are shown in Figure \ref{human-eval}. We can observe from the figure that our method generates answers that are more acceptable to humans compared to other baselines, maintaining a relatively high win rate in the competition. Only a small number of times does {\model} perform weaker than the baselines, and most of the time {\model} is equal or even better. Therefore, combining the above three different perspectives of evaluation, we can conclude that {\model} achieves outperforming results in the cloud product QA scenario.

%\begin{figure}[]
%  \centering
%\includegraphics[width=0.8\linewidth]%{pictures/heatmap.png}
%  \caption{Heat map for human evaluation. The %results}
%  \label{human-eval}
%\end{figure}
% \vspace{-8pt}
\subsection{Ablation Study (Q2)}
We conduct Ablation experiments to verify the validity of each module design. We validated the effectiveness of the designed components in our {\model}. We can find that the fine-tuning objective $\mathcal{L}_{ft}$ and the alignment objective $\mathcal{L}_{align}$ are both contributing to the model performance. Without fine-tuning (FT), the model performance can take a serious dip, as the LLM is not tuned to fit the golden answer. Besides, both two preference sets (SPS and KPS) in {\model} are contributing to the performance. The adaptive weights (AW) can control for the participation of different quality samples in the loss, which is also effective in {\model}.

\par Besides, we demonstrate the necessity of the CPKG with two groups of experiments. w/o RK denotes the experiment that removes the retrieved knowledge in the input prompt during the fine-tuning and preference alignment process. w/o KG denotes the experiment without KG in 
the whole process, which means the KPS and RK in the input prompt are all removed. For the results of these two groups of experiments, we can observe that the CPKG plays a remarkable role in {\model}. In the design of {\model}, the CPKG does not only serve as an external knowledge source during training but also participates in the preference set construction process, which is important to the model performance. In summary, each detailed design in our method {\model} has its unique role and contributes to the overall performance.
\begin{table}[]
\caption{The ablation study results. 
% We demonstrate the effectiveness of each module by removing it during experiments. 
We evaluate various stripped-down versions of our model to compare the performance gain brought by different components.
The full names of these abbreviations are as follows: FT (fine-tuning); AW (adaptive weight); SPS (style preference sets); KPS (knowledge preference sets); RK (retrieved knowledge).
}
\centering
\resizebox{0.88\linewidth}{!}{
\begin{tabular}{l|cccc}
\toprule
\multicolumn{1}{c|}{Setting} & \textbf{BLEU-1$\uparrow$} & \textbf{ROUGE-1}$\uparrow$ & \textbf{Reward}$\uparrow$ & \textbf{PPL}$\downarrow$   \\
\midrule
{\model}                      & 22.56  & 20.28   & -1.69  & 29.93 \\
\quad w/o FT                       & 13.17 & 12.91 & -2.14 & 31.96 \\
\quad w/o AW                      & 21.87  & 19.91   & -1.71  & 30.84 \\
\quad w/o SPS                      & 17.57  & 17.66   & -1.75  & 31.08 \\
\quad w/o KPS                      & 16.12  & 16.51   & -1.79  & 30.82 \\
\quad w/o RK                       & 17.46  & 17.56   & -1.89  & 30.85 \\
\quad w/o KG                       & 15.09 & 16.55 & -2.09 & 33.50 \\

% \quad w/o Align                    & 14.89 & 14.77 & -1.64 & 31.13 \\

\bottomrule
\end{tabular}
}
\end{table}

\subsection{Case Study (Q3)}
\begin{table}[]
\caption{The case study results for ground truth (GT), our {\model} predictions, and RRHF \cite{DBLP:journals/corr/rrhf} results. The original Chinese text have been translated into English for clarity.}
\label{case}
\centering
\resizebox{0.88\linewidth}{!}{
\begin{tabular}{l|l}
\toprule
\textbf{Question} & \begin{tabular}[c]{@{}l@{}}Please provide the steps for handling IOPS \\ detection errors.\end{tabular}                           \\
\midrule
GT     & \begin{tabular}[c]{@{}l@{}}It is recommended to replace the disk with one \\ that meets the IOPS specification.\end{tabular}      \\
\midrule
Ours        & \begin{tabular}[c]{@{}l@{}}It is recommended to replace the server with\\ device that meets the IOPS specifications.\end{tabular} \\
\midrule
RRHF       & \begin{tabular}[c]{@{}l@{}}After ADAC troubleshooting, restart the business \\ and check whether it is valid.\end{tabular} \\
\bottomrule
\end{tabular}
}
\resizebox{0.88\linewidth}{!}{
\begin{tabular}{l|l}
\toprule
\textbf{Question} & \begin{tabular}[c]{@{}l@{}}What is the explanation for the hwFlowRestoreFailed\\ alarm in CloudEngine 1800V product?\end{tabular}       \\
\midrule
GT     & \begin{tabular}[c]{@{}l@{}}The switch flow table restore failed (host\_ip=\\ {[}host\_ip{]}, host\_name={[}host\_name{]})\end{tabular}  \\
\midrule
Ours        & \begin{tabular}[c]{@{}l@{}}The switch flow table restore failed. (host\_ip=\\ {[}host\_ip{]}, host\_name={[}host\_name{]})\end{tabular} \\
\midrule
RRHF       & Flow table restore failed  \\
\bottomrule
\end{tabular}
}
\vspace{-16pt}
\end{table}

To make an intuition for the effectiveness of our method, we conduct a case study as shown in Table \ref{case}. We can observe that the answers generated by {\model} are more similar to the golden answer while keeping a user-friendly tone and providing sufficient information such as the host parameters in the second case. This suggests that the model learns appropriate style preferences. Besides, the retrieved knowledge in the first case is (EIP, used for, IP Binding), (Select Box, belongs to, Alarm Management Component), etc., which are all helpless to answer this question. However, {\model} is not misled by this useless knowledge and generates the correct answer while RRHF falls into the trap.

\subsection{Knowledge Retention Analysis (Q4)}

\begin{figure}[h]
  \centering
\includegraphics[width=0.9\linewidth]{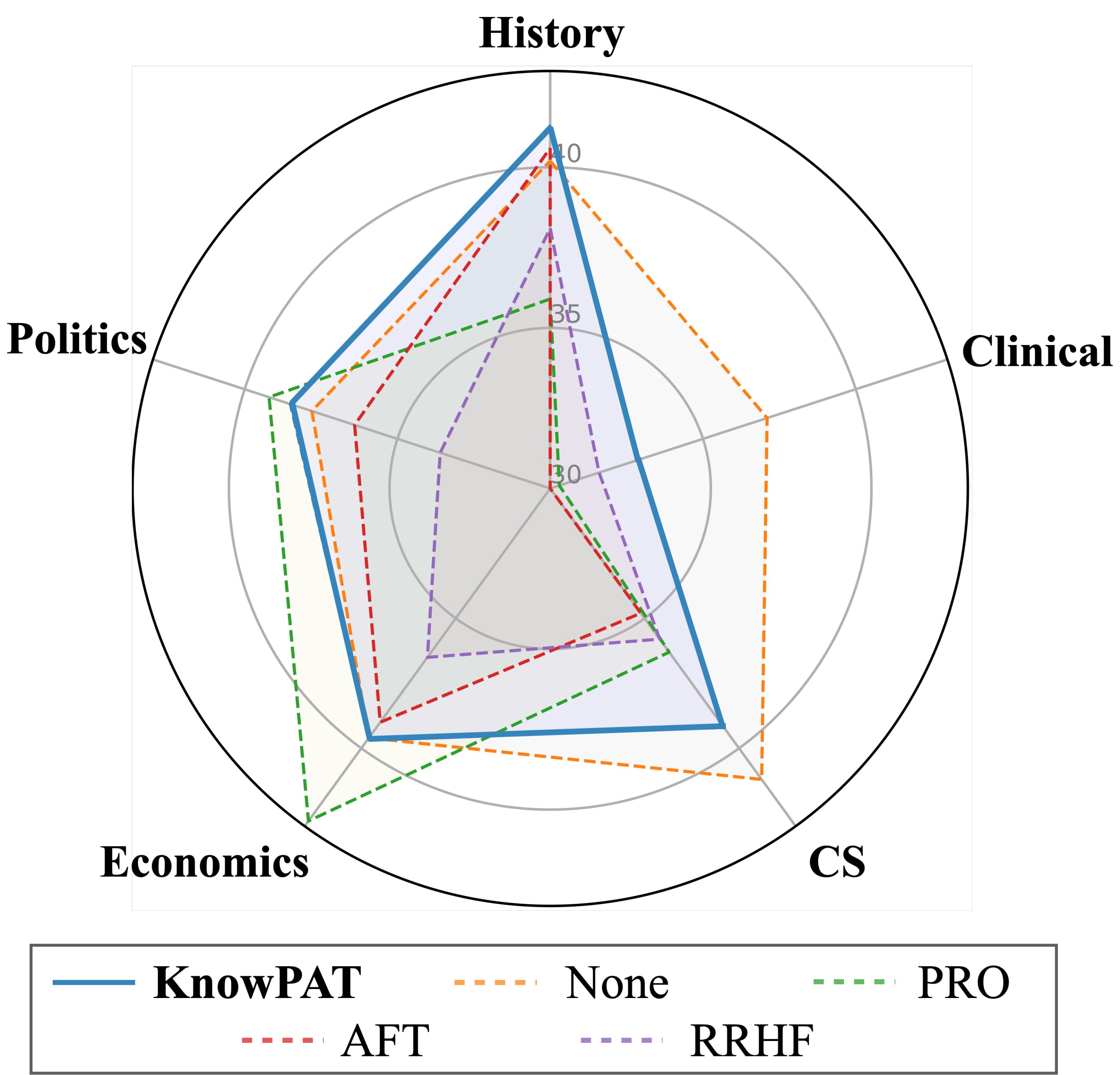}
  \caption{The commonsense ability on five domains.}
  \label{general-eval}
\end{figure}

% 改图，文字更靠近，去掉白边

As a project that needs to get off the ground in real-world scenarios, the general ability of the trained model should also be carefully evaluated, because the user may ask various kinds of questions if they like the model. We expect the model to keep their existing knowledge learned during pre-training and obtain new knowledge about our domain. Thus, we also conduct a commonsense evaluation on the trained models with the CMMLU \cite{li2023cmmlu} dataset, which is a benchmark for LLM's Chinese ability evaluation. The evaluation result is shown in \ref{general-eval}. We demonstrate the general ability on five distinctive commonsense regions (history, clinical, politics, computer science, and economics) for KnowPAT, vanilla Atom-7B (none), and other PA methods. As can be seen from the radargram, there is a relatively significant decline in the {\model}'s ability in medicine, but in the areas of politics, history, and economics it still maintains the ability of the original backbone model and even grows slightly. PRO, while unexpectedly showing a significant improvement in the economics problem, shows a more pronounced performance degradation than {\model} in several other areas.
Taken together, such variations of {\model} in generalized ability are acceptable for our cloud product QA scenario.

\section{Related Works}
\textbf{Preference alignment} (PA) \cite{DBLP:journals/corr/alignment, DBLP:journals/corr/abs-2309-03126-EPO} seeks to tailor pre-trained LLMs to align with human preferences (feedbacks) \cite{DBLP:conf/nips/instructgpt}. 
RLHF is a landmark work for PA, which leverages reinforcement learning (RL) \cite{DBLP:journals/corr/ppo} to align human preference with LLMs. Due to the sensitivity of RL parameters and the intricate three-stage processes of RLHF, many PA approaches have been proposed to address these challenges. For example, RRHF \cite{DBLP:journals/corr/rrhf} propose a margin-rank loss to optimize the LLMs without the need for extra reward models. PRO \cite{DBLP:journals/corr/pro} optimizes complex preference data with a list-wise contrastive loss. 
DPO \cite{DBLP:journals/corr/dpo} propose a direct preference optimization method by treating the LLM itself as a reward model. AFT \cite{DBLP:journals/corr/AFT} propose a ranking-feedback boundary-constrained alignment loss to optimize the preference data. Besides, our work also focuses on the large language model application and knowledge-enhanced QA. We give a brief introduction of these fields in Appendix \ref{appendix::related1} and \ref{appendix::related2}.
\section{Conlusion}
In this paper, we introduce a novel framework, knowledgeable preference alignment ({\model}), for domain-specific QA tasks in cloud product services, leveraging LLMs and KGs in a practical application setting. 
Our approach constructs a knowledgeable preference set by retrieving and utilizing knowledge triples to generate answers with different quantities. A new alignment objective is designed to unleash the power of the preference set.
Comprehensive experiments demonstrate that our method surpasses existing solutions for this real-world challenge.
Looking ahead, we aim to apply {\model} to more real scenarios such as enterprise-class services and further investigate the potential of KG-enhanced LLM application in the future.

\section*{Acknowledgment}

This work is founded by National Natural Science Foundation of China ( NSFC62306276 / NSFCU23B2055 / NSFCU19B2027 / NSFC91846204 ), Zhejiang Provincial Natural Science Foundation of China (No. LQ23F020017), Ningbo Natural Science Foundation (2023J291), Yongjiang Talent Introduction Programme (2022A-238-G),  Fundamental Research Funds for the Central Universities (226-2023-00138).

\section*{Limitations}
In this paper, we mainly focuses on a real-world application problem to align LLMs with knowledge preference for better domain-specific QA. There are still some limitations in our work. 

\noindent \textbf{Domain-specific scenario.}
Our approach is designed for specific domain (cloud product QA in our paper), and its effectiveness on general domains and open-source datasets is still subject to further validation. This will be the goal of our future endeavours.

\noindent \textbf{Forms of external knowledge.}
In our paper, we apply knowledge bases to store the external background knowledge for the QA tasks. This is a convenient and efficient way of storing knowledge for our scenario, but in more other scenarios, knowledge may be stored in other forms (e.g. unstructured text). Therefore, a more general framework to process the external knowledge with any format (KGs, unstructured text, documents) should be considered for better usage, which is also our future plan.
\section*{Ethical Considerations}
In this paper, we employ the open-source LLM to validate the effectiveness of our approach. Besides, the dataset we used is manually labeled with golden answer from domain experts engaged legally with suitable work intensity and well above aversage wages. Their rights are well protected at work. The content of the dataset is mainly questions about our cloud product usage, which do \textbf{not involve private information and sensitive data} of the target users. We promise that the content and collection steps of our dataset that are not against scientific ethics.

\bibliography{custom}

\appendix
\section*{Appendix}
\section{Related Works}
\subsection{KG-enhanced Question Answering}
\label{appendix::related1}
Knowledge graphs (KGs) \cite{DBLP:journals/tkde/kg-urvey, liang2022kgsurvey} is a kind of complex semantic web that models world knowledge in terms of structural triples as (\textit{head entity, relation, tail entity}). KGs {serve as } external knowledge source and benefit many AI tasks like language model pre-training \cite{DBLP:conf/aaai/k-bert}, question answering \cite{DBLP:conf/naacl/qa-gnn, wang2023vqagnn}, and recommendation systems \cite{DBLP:conf/kdd/KGAT, DBLP:conf/cikm/mmkg-rs}. Besides, domain-specific KGs are the important infrastructure of internet industry to provide exact factual knowledge, which is widely leveraged in E-commerce \cite{DBLP:conf/mm/mmkg-ecom, DBLP:conf/icde/mmkg-ecom2}, telecom fault analysis \cite{DBLP:conf/icde/telekg}, health care \cite{DBLP:journals/artmed/mmkghc1, DBLP:journals/ipm/mmkg-hc2} and so on. It is a popular topic to utilize KGs in real industry applications. 
In our scenario, we construct a domain-specific KG for cloud service products to benefit our Question Answering (QA) task.
{QA stands as a cornerstone in NLP, aiming at equipping machines with the capability to autonomously respond to human queries \cite{DBLP:conf/acl-mrqa/qa1, DBLP:conf/pkdd/qa2}.}
% QA tasks also come in a variety of forms. 
{QA tasks can take on various forms.}
% There are QA tasks that 
{Some require the selection from multiple choices, as seen in certain knowledge base QA (KBQA) \cite{DBLP:journals/pvldb/kbqa1, DBLP:journals/corr/kbqa2, DBLP:journals/corr/kbqa3} and visual question answering (VQA) \cite{DBLP:conf/iccv/vqa1, DBLP:conf/semweb/vqa2, DBLP:journals/pami/vqa3}.}
Conversely, tasks like open-domain QA often challenge systems to directly produce textual responses without a set answer pool \cite{DBLP:journals/tois/gqa1,DBLP:conf/emnlp/KarpukhinOMLWEC20}. In the last few years, fine-tuning pre-trained language models has been a leading approach for QA tasks. Models like BERT \cite{DBLP:conf/naacl/BERT} and T5 \cite{DBLP:journals/jmlr/T5} have previously achieved notable performance when adapted with question-answer pairs.

% \par 
% Besides, 
% QA is one of the main scenarios in which AI-related technologies land in real life. Many industry works tried to build automated domain-specific QA systems \cite{DBLP:journals/tois/gqa1,DBLP:conf/wsdm/gqa2} to serve their users. 
{We hold that QA doesn't just remain an academic pursuit; it acts as a bridge, facilitating the adoption of AI technologies in real-world applications. Numerous industrial efforts have been directed toward developing domain-specific QA systems to meet the needs of their users \cite{DBLP:journals/tois/gqa1,DBLP:conf/wsdm/gqa2}.}
% Domain-specific QA is often assisted by domain-specific knowledge bases (e.g. KGs), which can be retrieved to provide background knowledge for the input question. The task scenario we are facing is a question and answer in the vertical field of cloud service products, which is a domain-specific QA scenario.
{ Such systems often rely on domain-specific knowledge bases, like Knowledge Graphs (KGs), to provide relevant information for the posed questions. 
Our current investigation aligns with this trend, focusing on a domain-specific QA scenario for cloud service products.}
{Moreover, our approach diverges from these recent KG-based QA systems \cite{DBLP:conf/iclr/JiangZ0W23,DBLP:journals/corr/abs-2305-09645,DBLP:journals/corr/abs-2310-08975,DBLP:conf/acl/ZhangZWCHL0L23,DBLP:conf/jist/0007HCGFP0Z22} that utilize prompts for dialog with (large) language models to facilitate path reasoning and refine the scope of KG retrieval. We propose an innovative knowledgeable preference alignment framework that enhances KG-aware QA with the} knowledge preference.
% 我们提供了一种新的KG中知识融入LLM是范式，这种范式是检索-free的，灵活的，可以适配任何形式的KB，不需要针对地设计相关检索命令，在未来这有很重要的意义。可以直接把知识的偏好蒸馏/对齐到大模型的参数中，实现真正意义上的end-2-end
% 我们的方法是可迁移的，相似的paradigm可以直接引用到各种需要Knowledge的场景中。同时是可插拔，可以直接作用于任何模型进行领域知识注入。

\subsection{Large Language Model Application}
\label{appendix::related2}
Prominent large language models (LLMs) like GPT \cite{DBLP:journals/corr/gpt4,DBLP:conf/nips/gpt3,DBLP:conf/nips/instructgpt} and GLM \cite{DBLP:conf/iclr/chatglm,DBLP:conf/acl/glm} are sparking a wave of research in the community due to their generalization ability in many NLP tasks such as relation extraction \cite{DBLP:journals/corr/llm4kgc-zhu}, algebraic reasoning \cite{DBLP:conf/nips/llm4math}, and question answering \cite{llm4qa1, llm4qa2}. Most LLMs leverage the transformer \cite{DBLP:conf/nips/transformer} architecture, benefiting from training on vast corpora \cite{thakkar2023selfinfluence} through autoregressive tasks.
% {When adapting LLMs for specific downstream applications, supervised fine-tuning (SFT) is a prevalent methodology.}
Deploying and applying LLMs in real-life scenarios is also a major topic in industry today and several efforts have been made. For example, many works \cite{DBLP:conf/recsys/llm4rec1, DBLP:conf/recsys/llm4rec2, DBLP:journals/corr/llm4rec3, DBLP:journals/corr/llm4rec4, DBLP:conf/recsys/llm4rec5,DBLP:journals/corr/llm4rec6} attempt to build recommendation systems with LLMs. Some work like Huatuo \cite{wang2023huatuo} and LawyerLlama \cite{huang2023lawyer} have developed LLMs for domain-specific usage.

% Preference alignment (PA) \cite{DBLP:journals/corr/alignment,ji2023alignment-survey,DBLP:journals/corr/abs-2301-11259} seeks to tailor pre-trained LLMs to align with human preferences, be it social ethics, universal values, or specific linguistic styles, often termed as human feedback \cite{DBLP:conf/nips/instructgpt}.
% RLHF \cite{DBLP:conf/nips/instructgpt} is a landmark work for PA, which leverages reinforcement learning (RL) \cite{DBLP:journals/corr/ppo} to align the human preference with LLMs. Due to the sensitivity of RL parameters and the intricate three-stage structure of RLHF, many PA approaches \cite{DBLP:journals/corr/slic-hf, DBLP:journals/corr/dpo, DBLP:journals/corr/pro, nathani2023alignment} have been proposed to address these challenges. 

\par Our work proposes a knowledgeable preference alignment framework to incorporate the domain-specific KG into the preference alignment pipeline for the LLM application. By constructing a knowledgeable preference set, the LLMs are trained to align the knowledge preference with humans and select better factual knowledge in the input prompt to solve the QA task.
\section{Experiment Details}

\subsection{Dataset Details}
\label{appendix::dataset}
The detailed information of our dataset are shown in this section. We evaluate the performance of {\model} on two dataset: one private dataset CPQA constructed by us, and one public dataset RJUA-QA \cite{RJUA}. They are both domain-specific datasets.
\begin{itemize}
    \item CPQA is for the cloud product domain labeled by a team of human experts with 8909 QA pairs. CPQA employs a cloud product knowledge graph (CPKG) as the domain KB.
    \item RJUA-QA is a urological domain open-source dataset extracted from real-world medical records with 2132 QA pairs. RJUA-QA labels a series of medical context documents for each QA pair. We collect these contexts in the form of documents as the domain KB.
\end{itemize}

\subsection{Baseline Details}
\label{appendix::baseline}
\noindent \textbf{{(\rmnum{1})}} \textbf{Zero-shot approach}, which directly prompts the LLM with the input question to get the answer without training.

\noindent \textbf{{(\rmnum{2})}} \textbf{In-context learning \cite{DBLP:journals/corr/icl-survey} approach}, which would sample a few ($k$-shot) QA pairs as demonstrations from the training dataset as examples and get the answers from the LLM without training.

\noindent \textbf{{(\rmnum{3})}} \textbf{Vanilla fine-tuning approach}, which fine-tunes the LLM using the QA pairs w/ or w/o retrieved knowledge as Equation \ref{vft}. The fine-tuning baseline with retrieved knowledge is also known as retrieve-augmented generation (RAG) method.

\noindent \textbf{{(\rmnum{4})}} \textbf{Preference alignment approaches}, which introduce additional preference alignment objectives during training to align with human preference. We select five existing state-of-the-art (SOTA) PA methods including RRHF \cite{DBLP:journals/corr/rrhf}, SLiC-HF \cite{DBLP:journals/corr/slic-hf}, DPO \cite{DBLP:journals/corr/dpo}, PRO \cite{DBLP:journals/corr/pro}, AFT (both AFT-{BC} and AFT-{DC}) \cite{DBLP:journals/corr/AFT} as our baselines.

\subsection{Evaluation Details}
\label{appendix::evaluation}
We select three types of metrics to evaluate our method against baselines. The detailed information on the metrics is listed in the following:

\par \textbf{{(\rmnum{1})}} \textbf{Traditional text generation metrics.} We select several traditional text generation metrics such as BLEU \cite{DBLP:conf/acl/BLEU}, ROUGE \cite{lin2004rouge}, CIDEr \cite{DBLP:conf/cvpr/CIDEr}, and METEOR \cite{DBLP:conf/acl/METEOR} to evaluate the generated answers. However, these evaluation metrics are mainly used to measure the text-level similarity between generated answers and real answers, which means they can not fully reflect the semantic relevance or depth of understanding of the text.

\par \textbf{{(\rmnum{2})}} \textbf{Model-based metrics.} To evaluate the semantic similarity of the generated answers and the golden answers, we employ several model-based metrics such as BERTScore \cite{DBLP:conf/iclr/BERTSCORE}, perplexity (PPL), and preference score. These metrics evaluate the generated answers using various language models.
BERTScore employs BERT \cite{DBLP:conf/naacl/BERT} to calculate the semantic similarity between two sentences. PPL measures the ability of the LLM to understand and predict the entire sentence. The preference score is $\mathcal{S}$ mentioned in Equation \ref{score} to reflect the model's preference degree of the current answer.

\par \textbf{{(\rmnum{3})}} \textbf{Manual evaluation metrics.} We employ human labelers to evaluate the results from different methods. The labeler makes a judgment on two answers from unknown sources in a single-blind situation, chooses the better one, and counts the results. The comparison result in each turn is recorded as win/tie/lose.

\par The three main categories of metrics respond to a certain part of the result's characteristics at three levels: similarity at the textual level, similarity at the semantic level, and human preference.

\subsection{Implemention Details}
\label{appendix::implemention}
\par For zero-shot baselines, we select several different LLMs (ChatGPT \cite{DBLP:conf/nips/instructgpt}, ChatGLM-6B \cite{DBLP:conf/iclr/chatglm}, Baichuan-7B \footnote{\url{https://github.com/baichuan-inc/Baichuan-7B}}, Vicuna-7B \footnote{\url{https://github.com/lm-sys/FastChat}}, and Atom-7B-CP) for the zeros-shot approach. For in-context learning, we sample 1,2,4,8-shot QA pairs as demonstrations to support the input question. For the PA methods, we leverage the official code of RRHF \cite{DBLP:journals/corr/rrhf} and implement other PA methods (SLiC-HF \cite{DBLP:journals/corr/slic-hf}, PRO \cite{DBLP:journals/corr/pro}, AFT \cite{DBLP:journals/corr/AFT}) based on the code to reproduce the results on our preference dataset. The selection of hyper-parameters is based on the original paper. Atom-7B-CP is employed as the backbone model for all the baseline methods such as in-context learning, vanilla fine-tuning, and PA methods.

% BioASQ

\end{document}